\documentclass[lettersize,journal]{IEEEtran}
\usepackage{amsmath,amsfonts}
\usepackage{array}

\usepackage[caption=false, font=footnotesize, labelformat=empty]{subfig}
\usepackage{textcomp}
\usepackage{stfloats}
\usepackage{url}
\usepackage{verbatim}
\usepackage{graphicx}
\usepackage{cite}
\makeatletter
\def\bstctlcite{\@ifnextchar[{\@bstctlcite}{\@bstctlcite[@auxout]}}
\def\@bstctlcite[#1]#2{\@bsphack
  \@for\@citeb:=#2\do{%
    \edef\@citeb{\expandafter\@firstofone\@citeb}%
    \if@filesw\immediate\write\csname #1\endcsname{\string\citation{\@citeb}}\fi}%
  \@esphack}
\makeatother

\usepackage[ruled,vlined]{algorithm2e}

\usepackage{booktabs}
\usepackage{multirow}

\usepackage[T1]{fontenc}

\usepackage{calc} 

\usepackage[pdfstartview=XYZ,
bookmarks=true,
colorlinks=true,
linkcolor=blue,
urlcolor=blue,
citecolor=blue,
bookmarks=true,
linktocpage=true, 
hyperindex=true
]{hyperref}

\usepackage{orcidlink}

\begin{document}
\bstctlcite{IEEEexample:BSTcontrol} 

\title{Steganography in Game Actions}

\author{Ching-Chun Chang and Isao Echizen

\thanks{This work was supported in part by the Japan Society for the Promotion of Science (JSPS) under KAKENHI Grants (JP21H04907 and JP24H00732), and in part by the Japan Science and Technology Agency (JST) under CREST Grants (JPMJCR18A6 and JPMJCR20D3), AIP Acceleration Grant (JPMJCR24U3) and K Program Grant (JPMJKP24C2).
}
\thanks{C.-C. Chang and I. Echizen are with the Information and Society Research Division, National Institute of Informatics, Tokyo, Japan.
}
\thanks{Correspondence: C.-C. Chang (email: ccchang@nii.ac.jp)
}
}

\maketitle

\begin{abstract}
The exchange of messages has always carried with it the timeless challenge of secrecy. From whispers in shadows to the enigmatic notes written in the margins of history, humanity has long sought ways to convey thoughts that remain imperceptible to all but the chosen few. In the intricate patterns of imagery, the nuanced modulation of sound and the meticulous orchestration of language, the challenge of subliminal communication has been addressed in various forms of steganography. However, the field faces a fundamental paradox: as the art of concealment advances, so too does the science of revelation, leading to an ongoing evolutionary interplay. This study seeks to extend the boundaries of what is considered a viable steganographic medium. We explore a steganographic paradigm, in which hidden information is communicated through the episodes of multiple agents interacting with an environment. Each agent, acting as an encoder, learns a policy to disguise the very existence of hidden messages within actions seemingly directed toward innocent objectives. Meanwhile, an observer, serving as a decoder, learns to associate behavioural patterns with their respective agents despite their dynamic nature, thereby unveiling the hidden messages. The interactions of agents are governed by the framework of multi-agent reinforcement learning and shaped by feedback from the observer. This framework encapsulates a game-theoretic dilemma, wherein agents face decisions between cooperating to create distinguishable behavioural patterns or defecting to pursue individually optimal yet potentially overlapping episodic actions. As a proof of concept, we exemplify action steganography through the game of labyrinth, a navigation task where subliminal communication is concealed within the act of steering toward a destination. The stego-system has been systematically validated through experimental evaluations, assessing its distortion and capacity alongside its secrecy and robustness when subjected to simulated passive and active adversaries.
\end{abstract}

\section{Introduction}

\IEEEPARstart{S}{teganography} is the art and science of hiding information within non-suspicious media, concealing the very existence of the hidden message~\cite{4655281, 771065, 935180, Fridrich:2009aa}. Throughout history, the pursuit of secrecy in communication has been an enduring challenge, from conspiratorial whispers to cryptic codes, keeping secrets hidden from all but the chosen few. In contrast to cryptography, which obscures the content of a message through encryption, steganography seeks to disguise the presence of the message itself within a \textit{cover} medium, resulting in a \textit{stego} medium, and thereby circumventing the peril posed by codebreakers~\cite{668971}. The term originates from the Greek words \textit{steganos} meaning `covered' and \textit{graphia} meaning `writing'~\cite{10.1007/3-540-61996-8_27}. In the intricate patterns of imagery, the nuanced modulation of sound and the meticulous orchestration of words, the development of steganography has flourished in various forms, primarily centred on visual, auditory and linguistic media~\cite{10.1080/0161-119291866883, 10.1007/3-540-61996-8_48, 1511007, chang-clark-2014-practical, Zhu:2018aa, ziegler-etal-2019-neural}, as subtle alterations in these domains often remain imperceptible to human perceptual systems~\cite{5387237}.

However, the field of steganography faces an inherent challenge: as the art of covering and concealing information advances, so too does the science for uncovering and revealing it~\cite{10.1007/10719724_5, 1203220, 10.1145/1411328.1411349}. This ceaseless race manifests the transient nature of security in steganography, where breakthroughs in concealment are continually counterbalanced by the parallel evolution of detection mechanisms~\cite{1040098, 6081929, 6197267, 7444146}. This dynamic interplay is eloquently captured in the words of Horace: \textit{Time will bring to light whatever is hidden; it will cover up and conceal what is now shining in splendour.}

Venturing beyond the boundaries of traditional media explored in prior work, this study introduces a steganographic paradigm founded on behavioural media. We propose a stego-system that encodes messages as the episodes of multiple agents, and decodes them by identifying the source of each observed episode. An episode refers to a complete trajectory of states that starts from an initial state and ends at a terminal state. The system comprises multiple agents and an observer: each agent learns to complete a given task within the environment while ensuring its episode is uniquely identifiable, whereas the observer learns to distinguish between the episodes of different agents. In essence, action steganography can be formulated as a multi-agent reinforcement learning framework, where multiple agents operate within a shared environment, and their interactions are indirectly shaped by feedback from the observer.

In summary, we introduce a stego-system that automatically learns to encode and decode hidden information within the actions of artificial intelligence agents for covert communication. The concept of action steganography is formalised and integrated into the general framework of steganography. A game-theoretic perspective is offered to model the strategic nature of multi-agent interactions, where agents face the decision of whether to collaborate by generating unique episodes that facilitate steganographic communications, or to act independently, prioritising optimal episodes that may inadvertently overlap. As a proof of concept, we apply action steganography to the game of labyrinth, a navigation problem in which each agent steers towards a destination while encoding a message symbol into its episode. Although framed within the context of games, this paradigm extends far beyond entertainment or simulation. In a broader sense, a game refers to an environment where strategic interactions occur between rational players or decision-makers with either conflicting or aligned interests. The principles underlying the game of labyrinth closely mirror real-world applications of motion planning in unmanned vehicles, mobile robots and other autonomous systems. To validate the proposed stego-system, we conducted a series of experiments within the game of labyrinth, systematically evaluating its performance across distortion, capacity, secrecy and robustness, while accounting for both passive and active adversaries (eavesdroppers and intruders).

The remainder of this paper is organised as follows. Section~\ref{sec:game} outlines the key concepts in reinforcement learning that underpin this study. Section~\ref{sec:steganography} formalises action steganography and explores its game-theoretic foundations. Section~\ref{sec:labyrinth} exemplifies the stego-system through the game of labyrinth and provides guidance on its implementation. Section~\ref{sec:experiments} presents the experimental results, including visualisations of learnt policies and performance evaluations across distortion, capacity, secrecy (against eavesdroppers) and robustness (against intruders). Finally, Section~\ref{sec:conclusions} concludes the paper with a summary of research findings and potential future directions.

\section{Foundations of Agency}\label{sec:game}
A rational agent is a computational entity that perceives its environment through sensors and acts upon it through actuators, with the aim of maximising its payoff by making goal-oriented decisions, often informed by past experience and knowledge. Trial-and-error learning is a fundamental mechanism through which agents embody agency\textemdash the capacity to act purposefully and autonomously. The concept of trial-and-error learning was embedded in some of the earliest contemplations on artificial intelligence. Alan Turing proposed a design for a pleasure-pain system~\cite{10.1093/philmat/4.3.256}, a rudimentary learning mechanism inspired by the law of effect~\cite{Thorndike:1898aa}. This psychological principle suggests that behaviours followed by positive outcomes are reinforced, whereas those followed by negative outcomes are discouraged. The design envisioned a machine capable of making tentative choices in situations of uncertainty, exploring possible actions when it lacked a clear solution. In Turing’s words,
\begin{quote}
\textit{My contention is that machines can be constructed which will simulate the behaviour of the human mind very closely. They will make mistakes at times, and at times they may make new and very interesting statements, and on the whole the output of them will be worth attention to the same sort of extent as the output of a human mind.}
\end{quote}
This system was designed not only to mimic human-like decision-making but also to learn from its successes and failures. By reinforcing successful actions and discarding unsuccessful ones based on feedback, it provided an early conceptual framework for what we now recognise as reinforcement learning. This insight laid the foundation for machines to autonomously adapt their behaviour through mechanisms akin to reward and punishment.

\subsection{Game \& Policy}
A game is concerned with how an intelligent agent should take actions and receives feedback in a dynamic environment in order to maximise cumulative rewards. At its core, the interaction between an agent and its environment is modelled as a \textit{Markov decision process}~\cite{BELLMAN:1957aa}. It defines the state space, action space, transition dynamics and reward structure that collectively govern how the agent can interact with and explore the world. Formally, a Markov decision process is defined by a tuple $(\mathcal{S}, \mathcal{A}, T, R, \gamma)$, where: 
\begin{itemize}
	\item $\mathcal{S}$ is the set of all possible states,
	\item $\mathcal{A}$ is the set of actions available to the agent,
	\item $T(s' \mid s, a)$ defines the transition probabilities between states, and
	\item $R(s, a)$ specifies the reward received for taking action $a$ in state $s$.
\end{itemize}
A fundamental assumption of this framework is the Markov property, which states that `the future is independent of the past given the present'. This means that the next state depends only on the current state and action, but not on any prior states or actions. This memoryless property of a stochastic process simplifies decision-making, allowing the agent to make optimal actions based solely on the current state. At each time-step, the agent perceives the current state $s$ of the environment, selects an action $a$ from its action space, and then applies this action to the environment. The environment responds by transitioning to a new state $s'$ based on the agent's action as well as the underlying dynamics, and provides a reward $r$ as feedback to the agent. This reward indicates the immediate outcome of the agent's action, guiding the agent towards learning a policy $\pi$ that maximises cumulative rewards over time.

A policy $\pi$ defines the agent’s strategy for selecting actions in each state to maximise cumulative rewards. Formally, a policy is a function that maps states to a distribution over actions. Policies can be classified as either deterministic or stochastic~\cite{sutton1998reinforcement}. A deterministic policy specifies the action chosen in each state $s$, expressed as:
\begin{equation}
	\pi(s) = a .
\end{equation}
A stochastic policy specifies a probability distribution over actions in state $s$, expressed as:
\begin{equation}
	\pi(a \mid s) = \mathbb{P}(a \mid s).
\end{equation}
The agent's objective is to learn the optimal $\pi^{*}$, which selects actions in each state to maximise the cumulative reward over time. Throughout this paper, we assume a deterministic policy unless otherwise specified. This assumption simplifies notation by focusing on deterministic actions rather than stochastic distributions over actions.

\subsection{Principle of Optimality}
The foundation of optimal policies lies in \textit{Bellman’s principle of optimality}~\cite{BELLMAN:2010aa}, stated as follows:
\begin{quote}
\textit{An optimal policy has the property that whatever the initial state and initial decision are, the remaining decisions must constitute an optimal policy with regard to the state resulting from the first decision.}
\end{quote}
This implies that the solution to a complex, multi-step decision problem can be broken down into a sequence of simpler, one-step decisions, each leading optimally to the next state and ultimately to the goal. This recursive property is crucial for calculating the cumulative reward, the total sum of all future rewards an agent expects to receive starting from a given state. However, far-future rewards are often less valuable than immediate ones due to factors like uncertainty and delayed impact. To reflect this, we consider the \textit{cumulative discounted reward} for balancing the agent’s focus between short-term and long-term gains. It applies a discount factor $\gamma$ to future rewards, reducing their weight as they move further into the future. The cumulative discounted reward for state $s$ is defined as:
\begin{equation}
	G(s) = R(s, a) + \gamma R(s', a') + \gamma^{2} R(s'', a'') + \dots
\end{equation}
Consider a \textit{state-value function} $V_{\pi}(s)$ that represents the expected value of cumulative discounted reward $G(s)$ an agent will receive starting from state $s$ following policy $\pi$ thereafter. For each state, the value function helps the agent understand how valuable it is to be in that state with the aim of reaching the goal. Mathematically, for a given state $s$ subject to policy $\pi$, the state-value function $V_{\pi}(s)$ is defined by the Bellman expectation equation as:
\begin{equation}
\begin{split}
	V_{\pi}(s) &= \mathbb{E}_{\pi} \left[ G(s) \mid s \right] \\
	&= \mathbb{E}_{\pi} \left[ R(s, a) + \gamma G(s') \mid s \right] \\
	&= \mathbb{E}_{\pi} \left[ R(s, a) + \gamma V_{\pi}(s') \mid s \right] ,
\end{split}
\end{equation}
where $R(s, a)$ is the immediate reward obtained from taking action $a$ in state $s$, $V_{\pi}(s')$ is the expected value of future rewards under policy $\pi$, and $\gamma$ is the discount factor balancing immediate and future rewards. To determine the optimal way to act in each state, we aim to maximise the value function across all possible policies. The optimal state-value function $V^{*}(s)$ is thus defined as the maximum expected value achievable from state s across all possible policies:
\begin{equation}
	V^{*}(s) = \max_{\pi} V_{\pi} (s) .
\end{equation}
To achieve this maximum, an agent can evaluate actions directly to find the best action $a$ that yields the highest value in each state, as given by the Bellman optimality equation: 
\begin{equation}
	V^{*}(s) = \max_{a} \mathbb{E} \left[ R(s, a) + \gamma V^{*}(s') \mid s \right] .
\end{equation}
This equation encapsulates the idea that the optimal value of a state is the maximum expected return achievable by taking the best action in that state and then acting optimally thereafter.

\subsection{Dynamic Programming}
To find an optimal policy, we can use \textit{policy iteration} and \textit{value iteration}, two dynamic programming approaches that build on the Bellman's principle of optimality~\cite{howard1962dynamic}. Policy iteration is a cyclic process that alternates between policy evaluation and policy improvement. Policy evaluation computes the state-value function $V_{\pi}(s)$ for a given policy $\pi$ by iteratively applying the Bellman expectation equation until convergence: 
\begin{equation} 
\begin{split}
	V_{\pi}(s) &\leftarrow \mathbb{E}_{\pi} \left[ R(s, a) + \gamma V_{\pi}(s') \mid s \right] \\
	&= \sum_{s'} T(s' \mid s, a) \left[ R(s, a) + \gamma V_{\pi}(s') \right] .
\end{split}
\end{equation}
Once $V_{\pi}(s)$ converges, policy improvement updates the policy by selecting the action that maximises the expected return in each state, making it greedy with respect to $V_{\pi}(s)$:
\begin{equation} 
	\pi'(s) = \arg\max_a \sum_{s'} T(s' \mid s, a) \left[ R(s, a) + \gamma V_{\pi}(s') \right]. 
\end{equation}
By repeating these steps until the policy no longer changes, policy iteration converges to an optimal policy. Value iteration, on the other hand, combines policy evaluation and policy improvement in a single step. It directly applies the Bellman optimality equation to update the state-value function:
\begin{equation}
\begin{split}
	V(s) &\leftarrow \max_{a} \mathbb{E} \left[ R(s, a) + \gamma V(s') \mid s \right] \\
	&= \max_{a} \sum_{s'} T(s' \mid s, a) \left[ R(s, a) + \gamma V^{*}(s') \right].
\end{split}
\end{equation}
This iterative update process approximates $V(s)$ closer to the optimal value function $V^{*}(s)$. Once the value function converges, it can be used to derive an optimal policy by selecting actions that maximise the expected return for each state:
\begin{equation} 
\pi^{*}(s) = \arg\max_a \sum_{s'} T(s' \mid s, a) \left[ R(s, a) + \gamma V^{*}(s') \right]. 
\end{equation}
Both iteration methods require knowledge of the environment’s full transition and reward models, and involve updating every state in the state space during each iteration. This makes it computationally expensive and impractical for environments with large state spaces or unknown dynamics.

\subsection{Reinforcement Learning}
Reinforcement learning allows an agent to learn directly from interactions with the environment, receiving feedback in the form of rewards or penalties, without relying on prior knowledge of the environment's dynamics. One core method of reinforcement learning is \textit{temporal difference learning}~\cite{Sutton:1988aa}. It builds on Bellman’s optimality principle but updates the value function incrementally, step-by-step, as the agent experiences state transitions and rewards. This makes temporal difference learning \textit{model-free}, as it learns the value function through sampled experiences rather than requiring full knowledge of the environment’s dynamics or synchronous updates across the entire state space. 

Let us define the \textit{action-value function} $Q_{\pi}(s, a)$ as the expected cumulative discounted reward of taking action $a$ in state $s$ and following a policy $\pi$ thereafter~\cite{watkins1989learning}. The action-value function evaluates the value of specific state-action pairs, offering a more direct way of guiding action selection compared to the state-value function $V_{\pi}(s)$ alone. Mathematically, the action-value function for a given policy $\pi$ is defined by the Bellman expectation equation as:
\begin{equation}
Q_{\pi}(s, a) = \mathbb{E}_{\pi} \left[ R(s, a) + \gamma Q_{\pi}(s', a') \mid s, a \right] ,
\end{equation}
and the optimal action-value function can be defined by the Bellman optimality equation as
\begin{equation}
Q^{*}(s, a) = \mathbb{E} \left[ R(s, a) + \gamma \max_{a'} Q^{*}(s', a') \mid s, a \right] .
\end{equation}
Within the framework of temporal difference learning, on-policy and off-policy paradigms offer two primary approaches for learning value functions. The on-policy paradigm learns the action-value function based on the actions the agent actually takes under its current policy. For each visited state-action pair, the value is updated as
\begin{equation}
	Q_{\pi}(s, a) + \alpha \left[ R(s, a) + \gamma Q_{\pi}(s', a') - Q_{\pi}(s, a) \right],
\end{equation}
where $Q_{\pi}(s', a')$ is the value of the next state-action pair under the current policy $\pi$. In contrast, the off-policy paradigm aims to learn the optimal action-value function independent of the agent’s current policy. It approximates the optimal policy by assuming that the agent always takes the greedy action that maximises expected rewards. The value for each visited state-action pair is thus updated using the maximum future value as
\begin{equation}
    Q(s, a) + \alpha \left[ R(s, a) + \gamma \max_{a'} Q(s', a') - Q(s, a) \right],
\end{equation}
where $\max_{a'} Q(s', a')$ represents the maximum value for all possible actions $a'$ in the next state $s'$.

\subsection{Connectionism}
The tabular representations of $Q(s, a)$ are limited in that the value of each state-action pair must be explicitly stored. While these representations can be effective for environments with small and discrete state spaces, they become infeasible to manage in complex environments where the state space is large or continuous, due to the \textit{curse of dimensionality}. This limitation motivates the use of connectionist approaches, specifically \textit{artificial neural networks}~\cite{McCulloch:1943aa, Rosenblatt:1958aa, Hopfield:1982aa, 6796673, LeCun:2015aa}, which can generalise across states by learning underlying patterns and features. A seminal work in connectionist reinforcement learning was developed by DeepMind, known as the deep Q-network (DQN)~\cite{Mnih:2015aa}. The limitation is addressed by using an artificial neural network to approximate $Q(s, a)$ as a function parameterised by $\theta$, denoted as $Q_{\theta}(s, a)$, enabling the agent to operate in high-dimensional environments. 

The neural network is updated incrementally based on new experiences as the agent explores the environment. However, consecutive experiences are often highly correlated because they come from consecutive steps within the same episode. This correlation can lead to instability in the learning process, as it violates the assumption of independent and identically distributed (i.i.d.) samples typically required for stable learning. To address this, \textit{experience replay} was introduced to break up correlations by storing past experiences in an episodic buffer $\mathcal{B}$, allowing the agent to reuse and learn from a more diverse set of experiences~\cite{Lin:1992aa}. Let us denote an experience by a tuple $(s, a, r, s')$, representing a single interaction between the agent and the environment, where $s$ is the current state, $a$ is the action taken by the agent in the current state, $r$ is the immediate reward the agent receives after taking action $a$ in state $s$, and $s'$ is the next state the agent transitions to after taking action $a$ in state $s$. Given a collection of sampled experiences, the learning objective is to minimise the mean-squared error between the target values and the values predicted by the neural network, expressed as:
\begin{equation}
    \mathcal{L}(\theta) = \mathbb{E}_{(s, a, r, s')\sim \mathcal{B}} \left[ (y - Q_{\theta}(s, a))^2 \right],
\end{equation}
where the target value is given by:
\begin{equation}
    y = r + \gamma \max_{a'} Q_{\theta}(s', a').
\end{equation}
To further stabilise the learning process and reducing oscillations, the target value can be computed using a separate target neural network that maintains a fixed set of weights, periodically copied from the primary neural network. This prevents the primary neural network from `chasing its own predictions'.

\begin{figure*}[!t]
\centerline{\includegraphics[width=2.0\columnwidth]{Figures/overview.pdf}}
\caption{Overview of action steganography: Alice encodes a message via the interactions of an agent with an environment, whereas Bob decodes the message from the consequent episode.}
\label{fig:overview}
\end{figure*}

\section{Action Steganography}\label{sec:steganography}
The core challenge of steganography is to transmit information without raising suspicion, even under scrutiny. This challenge is illustrated by Gustavus Simmons’ problem of prisoners~\cite{Simmons1984}:
\begin{quote}
	\textit{Two accomplices in a crime have been arrested and are about to be locked in widely separated cells. Their only means of communication after they are locked up will be by way of messages conveyed for them by trustees—who are known to be agents of the warden.}
\end{quote}
This scenario embodies the fundamental principle of steganography: the prisoners must devise a method to communicate covertly without revealing the presence of their messages to the warden. In this study, we propose a steganographic communication via the behavioural patterns of agents interacting with an environment. The agents act as encoders, embedding a message in their distinct sequences of actions as they execute policies designed to accomplish a specific task. The objective is for these behavioural patterns to be recognisable by an observer, thereby allowing the message within the observed actions to be decoded. An overview of action steganography is illustrated in Figure~\ref{fig:overview}.

\subsection{Elements of Steganography}
In a typical steganographic communication scenario, a \textit{sender} (referred to as Alice $\mathfrak{A}$) wishes to send a hidden message $m$ to a \textit{receiver} (referred to as Bob $\mathfrak{B}$) without arousing suspicion. Let $\mathcal{M}$ denote the set of possible messages, $\mathcal{C}$ the set of possible cover media, and $\mathcal{S}$ the set of stego media. Alice encodes $m$ into a chosen cover medium $c$ (such as an image, audio, video or text) using an encoder function $E$, generating the stego medium $s$ that is statistically indistinguishable from the distribution of cover media, as given by:
\begin{equation}
s = E(m, c).
\end{equation}
The encoder function $E$ can be defined as
    \begin{equation}
    E : \mathcal{M} \times \mathcal{C} \rightarrow \mathcal{S}.
    \end{equation}
Upon receiving the stego medium $s$, Bob applies a decoder function $D$ to extract the message
\begin{equation}
\hat{m} = D(s).
\end{equation}
If decoding is successful, then $\hat{m} = m$; otherwise, $\hat{m} \neq m$. The decoder is defined as
    \begin{equation}
    D : \mathcal{S} \rightarrow \mathcal{M}.
    \end{equation}
 
Throughout this steganographic communication process, Alice and Bob must consider potential adversaries~\cite{10.1007/10719724_1, CACHIN200441, 4663056}. A passive adversary or \textit{eavesdropper} (referred to as Eve $\mathfrak{E}$) inspects the communication between Alice and Bob without interacting with the medium. Eve analyses the stego medium $s$ to detect if it contains hidden information. A passive adversarial function can be defined as
   \begin{equation}
   f_{\mathfrak{E}} : \mathcal{S} \rightarrow \{0, 1\},
   \end{equation}
   where $f_{\mathfrak{E}}(s) = 1$ indicates the detection of hidden content, and $f_{\mathfrak{E}}(s) = 0$ suggests no hidden message is found. An active adversary or \textit{intruder} (referred to as Trudy $\mathfrak{T}$) not only observes the communication but also intercepts and interferes with the stego medium to prevent successful decoding. Trudy may alter $s$ with the goal of corrupting the integrity of the hidden message, thereby preventing $m$ from being accurately decoded. An active adversarial function $f_{\mathfrak{T}}$ can be defined as
   \begin{equation}
   f_{\mathfrak{T}} : \mathcal{S} \times \mathcal{N} \rightarrow \mathcal{S},
   \end{equation}
   where $\mathcal{N}$ represents a set of noises or perturbations applied to $s$. The output $\tilde{s} = f_{\mathfrak{T}}(s, n)$ is an altered version of the stego medium with noise $n$ that is intended to disrupt the decoding process, so that $D(\tilde{s}) = \hat{m} \neq m$.

\subsection{Agent \& Observer}
Given an environment where multiple agents and an observer interact, we formulate a steganography framework as follows. Let $\mathcal{K}$ be a stego-key, encapsulating a shared collective configuration of hyper-parameters and random seeds between Alice and Bob. With the use of $\mathcal{K}$, Alice and Bob initialise their agents and observer in the given environment $\mathcal{V}$, where Alice has her $n$ agents $\{\mathcal{A}^\mathfrak{A}_{i} \}_{i=0}^{n}$ along with an observer $\mathcal{O}^\mathfrak{A}$, and Bob also has $n$ agents $\{\mathcal{A}^\mathfrak{B}_{i} \}_{i=0}^{n}$ along with an observer $\mathcal{O}^\mathfrak{B}$. Since $\mathcal{K}$ is shared between Alice and Bob, the resultant agents and observer should be identical; therefore, we may refer to the shared agents and observer as $\{\mathcal{A}_{i} \}_{i=0}^{n}$ and $\mathcal{O}$ for simplicity. In the context of an environment $\mathcal{V}$, Alice generates an episode $\boldsymbol{\epsilon}_{i}$ with an agent $\mathcal{A}_i$, which is a sequence of states ending at a terminal state or condition, represented as
\begin{equation}
	\boldsymbol{\epsilon}_{i} = \{s_{t}\}_{t=0}^{T_i},
\end{equation}
where the episode terminates at the final time-step $T_i$. This episode results from the feedback loop between the agent and the environment. For a state at time $t$, the agent takes an action
\begin{equation}
	a_{t} = \pi_{\mathcal{A}_i}(s_t),
\end{equation}
and the environment updates the state based on the current action as 
\begin{equation}
	s_{t+1} = \mathcal{V}(a_{t}) .
\end{equation}
At the other end, Bob deduces the identity of the agent from the episode $\boldsymbol{\epsilon}_{i}$ using the observer $\mathcal{O}$, represented as
\begin{equation}
	\hat{i} = \pi_{\mathcal{O}}(\boldsymbol{\epsilon}_{i}) .
\end{equation}

\subsection{Eavesdropper \& Intruder}
In the context of the steganographic framework illustrated in the provided setup, Alice and Bob aim to communicate covertly through the actions of multiple agents, while potential adversaries, Eve (the eavesdropper) and Trudy (the intruder), pose distinct threats to the security and integrity of this communication. We assume that the adversaries possess complete knowledge of the system, limiting the protection of the system solely to the secrecy of the key. This assumption follows Kerckhoffs's principle, phrased by Shannon as `\textit{the enemy knows the system}'~\cite{6769090}. In other words, a steganographic system should remain secure even if everything about the system—except the key—is known to the adversary.

Eve is a passive adversary who attempts to detect the presence of hidden information within the communication between Alice and Bob without directly interfering with the communication. She analyses the episode $\boldsymbol{\epsilon}_i$ generated by Alice’s agent $\mathcal{A}_i$ as it interacts with the environment $\mathcal{V}$, looking for clues such as behavioural patterns or statistical anomalies that may indicate subliminal communication. While the stego-key $\mathcal{K}$ is assumed to be kept secret between Alice and Bob, Eve can configure her own parameters to create shadow agents and build a detection mechanism to distinguish steganographic episodes from normal ones. Eve's objective is to maximise the probability of correctly classifying the type of episodes (either stego or cover) given an observed episode $\boldsymbol{\epsilon}_{\text{query}}$, optionally along with the sets of stego-episode $\mathcal{E}_{\text{stego}}$ and cover-episode $\mathcal{E}_{\text{cover}}$, as given by:
\begin{equation}
\max P( \hat{y} = y \mid \boldsymbol{\epsilon}_{\text{query}}, \mathcal{E}_{\text{stego}}, \mathcal{E}_{\text{cover}})),
\end{equation}
where $y$ denotes the true type of $\boldsymbol{\epsilon}_{\text{query}}$ and $\hat{y}$ denotes Eve's prediction.

Trudy is an active adversary who not only observes the communication but also seeks to actively interfere with it. He attempts to disrupt the covert channel by intercepting and deviating the stego episode $\boldsymbol{\epsilon}$ to prevent Alice and Bob from successfully transmitting the hidden message. His interference may distort the integrity of actions generated by Alice’s agent to an extent that effectively corrupts the intended information before it reaches Bob, while keeping the distortion within limits to avoid severely compromising the optimality of the actions. Trudy’s objective is to maximise the probability of erroneous message decoding, as given by:
\begin{equation}
\max P(\pi_{\mathcal{O}}(\tilde{\boldsymbol{\epsilon}_{i}}) \neq i) ,
\end{equation}
where $\tilde{\boldsymbol{\epsilon}_{i}}$ is a deviated episode with constrained interference.

\subsection{Game-Theoretic Equilibria}
Action steganography, where agents encode messages into trajectories while an observer attempts to decode the messages from these trajectories, is essentially a multi-agent reinforcement learning problem~\cite{10.5555/3091574.3091594, 10.5555/295240.295800, 10.5555/3295222.3295385}. It can be analysed from a game-theoretic perspective, particularly through the lens of the \textit{stag hunt dilemma}, which originates from philosopher Jean-Jacques Rousseau's \textit{Discourse on Inequality}. The stag hunt dilemma is a fundamental problem in game theory that illustrates the interplay between individual rationality and collective cooperation. It is an allegory that involves two hunters who must decide whether to cooperate in hunting a stag, which requires mutual effort, or defect by hunting a hare, which is achievable individually but yields a lower reward. The dilemma illustrates the tension between the higher potential payoff of \textit{cooperation} and the safer but less rewarding option of \textit{defection}. The payoffs in the game are shown in Table~\ref{tab:stag_hunt_payoffs} and defined as follows:
\begin{itemize}
	\item Optimism $O$: The payoff for mutual cooperation.
	\item Egoism $E$: The payoff for a player who defects while the other cooperates.	
	\item Pessimism $P$: The payoff for mutual defection.
	\item Altruism $A$: The payoff for a player who cooperates while the other defects.\end{itemize}
The relationships between these payoffs are critical to the dynamics of the stag hunt and can be expressed as:
\begin{equation}
	O > E \geq P > A.
\end{equation}
The optimism payoff is greater than the egoism payoff, emphasising the idea that hunting the stag together is more valuable than hunting hares alone, even though it requires coordination and trust. The egoism payoff is at least as good as the pessimism payoff, reflecting that solitary hare hunting can sometimes be more efficient than sharing resources, as competition for limited resources diminishes returns; equality holds in environments with abundant resources. The pessimism payoff is better than the altruism payoff, indicating the risk associated with cooperation: if one player defects, the cooperating player is left empty-handed, as the stag cannot be hunted alone. The payoff relationships highlight the dynamics of cooperation $C$ and defection $D$ in the stag hunt:
\begin{itemize}
	\item Mutual Cooperation $(C, C)$: This is the payoff-dominant Nash equilibrium and also the Pareto optimum, where both players trust each other to cooperate and achieve the maximum reward.
	\item Mutual Defection $(D, D)$: This is the risk-dominant Nash equilibrium, as it avoids the risk of one player being left vulnerable to unreciprocated cooperation.
	\item Mixed Strategies $(C, D)$ or $(D, C)$: These strategies are unstable as the cooperator suffers the altruism payoff, while the defector gains the egoism payoff.
\end{itemize}
A strategy profile is a Nash equilibrium if no rational player can unilaterally deviate and improve their payoff, given the strategies of the other players~\cite{Nash:1951aa}. A strategy profile is Pareto optimum if there is no other profile that makes at least one player better off without making any other player worse off~\cite{Arrow:1954aa}. In the stag hunt game, the interplay of cooperation and defection gives rise to two Nash equilibria: one that is payoff-dominant and one that is risk-dominant, reflecting the trade-offs between maximising mutual rewards and minimising individual risks, respectively. Furthermore, the Pareto optimum reflects the scenario where players achieve the highest possible collective payoff.

\begin{table}[!t]
\centering
\caption{Payoff matrix for the stag hunt dilemma.}
\resizebox{0.9\columnwidth}{!}{
\begin{tabular}{c|c|c}
\toprule
\midrule
\textbf{Players (A, B)} & \textbf{Cooperation} & \textbf{Defection} \\
\midrule
\textbf{Cooperation} & $(O, O)$ & $(A, E)$ \\
\midrule
\textbf{Defection} & $(E, A)$ & $(P, P)$ \\
\midrule
\bottomrule
\end{tabular}
}
\label{tab:stag_hunt_payoffs}
\end{table}

This framework models the strategic interplay between agents deciding whether to cooperate by creating distinguishable trajectories or to defect by prioritising optimal, potentially overlapping trajectories. The observer, in this case, is treated as a stationary \textit{oracle machine}~\cite{Turing:1939aa}, whose ability to decode the agents’ identities or messages depends on the distinguishability of their trajectories. A simple way to encourage an optimal solution is to set up an incentive structure that alleviates the fear of betrayal, which often deters agents from attempting cooperation. For instance, let the optimism payoff (mutual cooperation) be inherently more attractive than the pessimism payoff (mutual defection), and let the egoism payoff (unilateral defection) not be much higher than the altruism payoff (unilateral cooperation), thereby reducing the incentive for unilateral or mutual defection. Over time, the agents develop policies that reflect the trade-offs between cooperation and defection, leading to an equilibrium that balances trajectory efficiency and identifiability.

\section{Game of Labyrinth}\label{sec:labyrinth}
The labyrinth game exemplifies an ideal environment for demonstrating the fundamental mechanics of reinforcement learning within the steganographic framework. Claude Shannon's Theseus, a labyrinth-solving electromechanical mouse, serves as a precursor to modern concepts in artificial intelligence and was described as~\cite{shannon_macy}:
\begin{quote}
	\textit{A maze-solving machine that is capable of solving a maze by trial-and-error means, of remembering the solution, and also of forgetting it in case the situation changes and the solution is no longer applicable. I think this machine may be of interest in view of its connection with the problems of trial-and-error learning, forgetting, and feedback systems.}
\end{quote}
This machine was capable of navigating a labyrinth through trial-and-error exploration, learning the correct trajectory to the goal by systematically eliminating dead ends and storing the solution in its memory. This adaptive behaviour closely mirrors the principles of reinforcement learning, where an agent interacts with an environment, evaluates the outcomes of its actions, and improves its performance over time by maximising cumulative rewards.

\subsection{Definitions of Labyrinth}
Consider an environment in the form of a labyrinth. In labyrinth-solving or motion-planning game, the agent learns an optimal policy to navigate from start to goal with minimal steps while avoiding obstacles. Each component of this environment is defined as follows:
\begin{itemize}
	\item The \textit{state space} represents all possible positions within the labyrinth that the agent can occupy, with each state encoding the positions of the start, the goal, the obstacles and the agent itself.
	\item The \textit{action space} consists of all possible moves the agent can make, which are typically restricted to four primary directions: up, down, left and right, subject to the constraints imposed by the boundaries and obstacles.
	\item The \textit{transition function} in a deterministic labyrinth environment is straightforward, as taking an action from one state reliably results in a new state, unless an obstacle or boundary prevents movement, in which case the agent remains in the current state.
	\item The \textit{reward function} provides feedback by assigning a high positive reward upon reaching the goal to reinforce task completion, a small negative reward for each step to encourage shortest path, and additional penalties for hitting obstacles, crossing boundaries or retracing steps to discourage inefficient paths.
\end{itemize}

\subsection{Cellular Automata}
Cellular automation can serve as a method for labyrinth generation. Cellular automata, introduced by John von Neumann, were originally conceived as a theoretical framework to explore self-replicating systems~\cite{10.5555/1102024}. A cellular automaton consists of a grid of cells, each transitioning between states based on the states of its neighbours and a set of predetermined rules. This concept was inspired by the biological processes of reproduction and sought to understand how complex structures could arise from simple, local interactions. This idea was later popularised by Conway’s game of life~\cite{Gardner:1970aa}, a two-dimensional cellular automaton where each cell's state (alive or dead) evolves according to the following rules:
\begin{itemize}
	\item Underpopulation: Any live cell with fewer than two live neighbours dies.
	\item Overpopulation: Any live cell with more than three live neighbours dies.
	\item Reproduction: Any dead cell with exactly three live neighbours becomes a live cell.
\end{itemize}
To adapt cellular automata from the game of life for labyrinth generation, cells are assigned one of two states: pathway or obstacle. The process begins with a grid where each cell is randomly initialised as either a path or an obstacle with a given probability. To count neighbouring obstacles, we define a cell's neighbours by the Moore neighbourhood (8 surrounding cells). Let the thresholds for underpopulation, overpopulation and reproduction be denoted as $\theta_{\text{under}}$, $\theta_{\text{over}}$ and $\theta_{\text{re}}$, respectively. If the current cell at coordinate $(i, j)$ is an obstacle $c_{ij} = 1$, its state transitions by applying the underpopulation and overpopulation rules:
\begin{equation}
c'_{ij} =
\begin{cases}
1 & \text{if } (\theta_{\text{under}} \leq \operatorname{neighbour}_{ij} \leq \theta_{\text{over}}), \\
0 & \text{otherwise}.
\end{cases}
\end{equation}
If the current cell is a path $c_{ij}(t) = 0$, its state transitions by applying the reproduction rule:
\begin{equation}
c'_{ij} =
\begin{cases}
1 & \text{if } \operatorname{neighbour}_{ij} = \theta_{\text{re}}, \\
0 & \text{otherwise}.
\end{cases}
\end{equation}
To further increase diversity and stochasticity in labyrinth generation, we may introduce random transition probabilities for state changes, allowing transitions from pathway to obstacle and from obstacle to pathway to occur probabilistically rather than deterministically.

\subsection{Optimal Trajectory}
To identify the optimal trajectory in a labyrinth, we use \textit{Dijkstra’s algorithm}, a classic method in graph theory for finding the shortest path in a weighted graph~\cite{Dijkstra:1959aa}. This algorithm computes the minimum-cost path from a starting point, known as the source node, to other nodes in a graph. The labyrinth is represented as a graph $G = (V, E)$, where $V$ is the set of vertices (cells in the labyrinth) and $E$ is the set of edges (connections between neighbouring cells). Each edge has a weight, representing the cost of moving between two nodes. In the context of a labyrinth, these weights are usually uniform, as each step has an equal cost. Dijkstra’s algorithm begins by assigning a tentative distance $d(v)$ to every node $v$ in the graph. The source node is initialised with a distance of zero, while all other nodes are assigned a distance of infinity, indicating that they are initially unreachable. The algorithm maintains a priority queue, which always processes the node with the smallest tentative distance. At each step, the node $u$ with the smallest tentative distance is dequeued from the priority queue. The algorithm then examines the current node’s neighbours, updating the tentative distance of each neighbour $v$ if a shorter path is found through the current node. Mathematically, this update process, known as relaxation, is expressed as:
\begin{equation}
d(v) = \min(d(v), d(u) + w(u, v)),
\end{equation}
where $d(u)$ is the current distance to the neighbouring node $u$, and $w(u, v)$ is the weight of the edge connecting $u$ and $v$. This process continues until all nodes have been visited or until the shortest path to a specific goal node has been determined. The result is a set of shortest distances from the source to all reachable nodes, along with the paths taken to achieve them. The principle of Dijkstra’s algorithm is to use a greedy approach to iteratively expand the shortest known path from the source node, relying on the monotonicity of non-negative weights to ensure that once a node's shortest path is finalised (when the node is de-queued), no alternative path through other nodes can yield a shorter distance.

\subsection{Spatiotemporal Learning Machinery}
We now turn our attention to machine learning models for solving the game of labyrinth in the context of steganography. The state at each discrete time-step is represented as a multi-channel, one-hot encoded matrix that captures spatial relationships within this grid-based environment. This structured representation consists of distinct channels, each corresponding to a specific component within the environment: the agent, the goal, the obstacles. Each channel is represented by a binary matrix, where a value of $1$ in a cell indicates the presence of the respective element at that location, while $0$ denotes absence. The trajectory is a temporal sequence of multi-channel one-hot encoded states. The data representations inform the design of the following neural networks.

\begin{itemize}
	\item Agent's Neural Network: For the Q-function approximator (deep Q-network) of each agent, we construct a \textit{convolutional neural network} (CNN) architecture tailored for grid-based state representations~\cite{726791}. This model leverages convolutional layers to capture spatial features from an input state representation, followed by a fully connected layer that outputs Q-values for each possible action.
	
	\item Observer's Neural Network:	For the episodic classification model of the observer, we build a \textit{recurrent neural network} (RNN) architecture suited to episodic trajectories. This model utilises multi-layer bidirectional \textit{long short-term memory} (LSTM) modules to capture both forward and backward temporal dependencies across sequences of states~\cite{Hochreiter:1997aa}, followed by a fully connected layer that outputs a probability distribution over agent identities.
\end{itemize}
For each agent's neural network, the learning process begins by sampling a batch of experiences from the replay buffer, each consisting of a current state $s$, an action $a$, a reward $r$ and a next state $s'$ for each piece of experience. The estimated Q-value is obtained by passing the current state $s$ through the model and selecting the Q-value corresponding to the taken action $a$. The target Q-value is calculated using the Bellman equation, which combines the observed reward $r$ with the discounted maximum Q-value of the next state $s'$, selected from the Q-values for all possible actions obtained by passing the next state through the model. The loss, defined as the mean squared error between the current and target Q-values, is back-propagated to update the network weights using an optimiser. For the observer's neural network, the cross-entropy loss function is employed to measure the discrepancy between the predicted probabilities and the true identity labels. The predictions are obtained by passing the trajectories, represented as sequences of states, through the model.

\subsection{Feedback Structure}
With the learning framework established, we next delve into the feedback structure that governs rewards and penalties in reinforcement learning. This structure plays a pivotal role in shaping the agents' trajectories, balancing the objectives of efficient navigation and unambiguous identifiability simultaneously. The following reward and penalty components are designed to incentivise desirable behaviours while discouraging ineffective actions.
\begin{itemize}
	\item Time Penalty: Negative feedback is given for each step taken, which discourages the agent from taking unnecessarily long paths.
	\item Revisit Penalty: Negative feedback is applied if the agent revisits the same state multiple times, which discourages looping and inefficient paths.
	\item Collision Penalty: Negative feedback is given if the agent attempts to move into an obstacle or across a boundary.
	\item Terminal Reward: Positive feedback is given when the agent reaches the goal, reinforcing successful completion of navigation.
	\item Steganographic Reward: Feedback, whether positive or negative, depends on the identifiability of the agent's episode by the observer.
\end{itemize}

\begin{figure}[!t]
\centerline{\includegraphics[width=0.73\columnwidth]{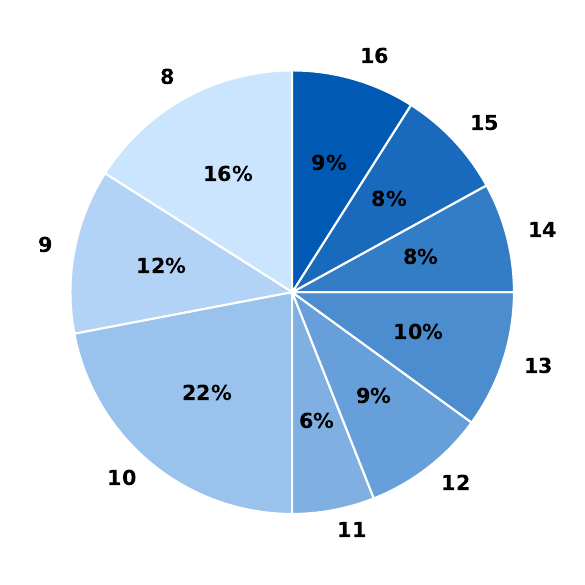}}
\caption{Proportion of labyrinths categorised by obstacle count.}
\label{fig:obstacle_count}
\end{figure}

\subsection{Exploration-Exploitation Dilemma}
In reinforcement learning, an agent needs to balance between exploring new actions and exploiting learnt knowledge, referred to as the \textit{exploration-exploitation dilemma}. Exploration allows the agent to discover potentially better actions that may lead to higher cumulative rewards, while exploitation serves to reinforce the agent’s understanding of successful actions. Balancing these two objectives is essential for the agent to avoid getting stuck in suboptimal behaviours and to fully explore the environment. During the learning phase, we may follow a stochastic \textit{Boltzmann policy}, which samples an action from a Boltzmann distribution with the probability of each action being proportional to its Q-value:
\begin{equation}
	a_{t} \sim \pi(a \mid s_t) = \frac{\exp (Q_{\pi}(s_t, a)/\tau)}{\sum_{b} \exp (Q_{\pi}(s_t, b)/\tau)} ,
\end{equation}
where $\tau$ is the temperature parameter that controls the degree of exploration. A higher value of $\tau$ results in more uniform probabilities across actions, encouraging exploration. Conversely, a lower value of $\tau$ sharpens the focus on actions with higher Q-values, favouring exploitation. Annealing the temperature over time allows the agent to explore broadly in the early stages, while slowly shifting toward exploitation as it gains confidence in its learned policy. Additionally, we may allow the agent to select an equiprobable random action with a small probability, mitigating the risk of premature convergence to suboptimal policies and enhancing adaptability in complex or stochastic environments. During inference, the agent follows a deterministic greedy policy, selecting actions based solely on maximising the estimated Q-values:
\begin{equation}
	a_{t} = \pi(s_t) = \arg \max_{a} Q_{\pi}(s_t, a) .
\end{equation}

\begin{figure*}[!t]
\centerline{\includegraphics[width=2.0\columnwidth]{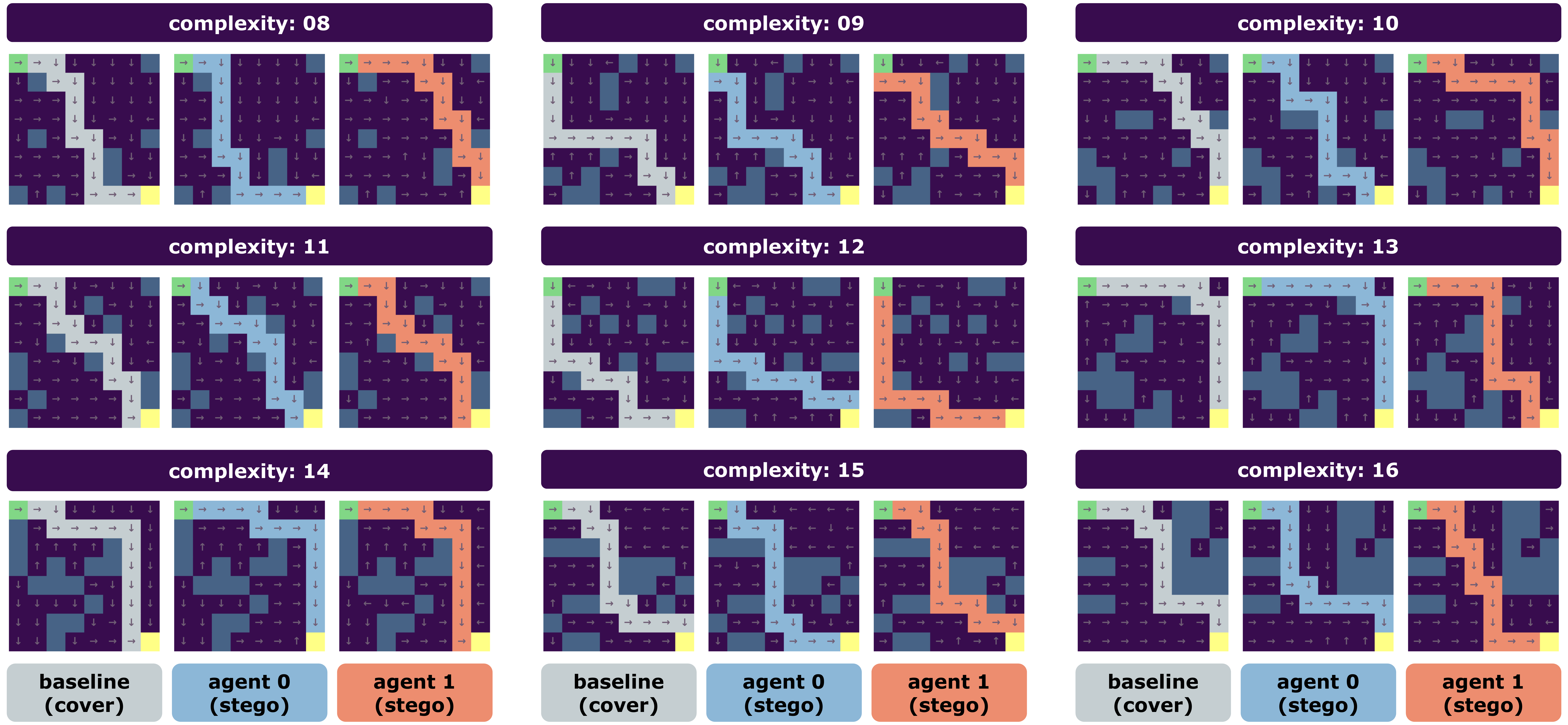}}
\caption{Visualisation of cover and stego policies across labyrinths of varying complexity.}
\label{fig:vis_maze}
\end{figure*}

\section{Experiments}\label{sec:experiments}
This section details the experiments conducted to evaluate the proposed action steganography. It begins with the experimental setup and visualisation of the agents' learnt policies, followed by a quantitative assessment of the distortion introduced by steganographic communication and the capacity of the steganographic channel. Finally, the secrecy and robustness of the stego-system are analysed through evaluations against simulated adversaries.

\subsection{Experimental Setup}
The experimental setup, covering the game environment, feedback structure, model architectures and learning process, is laid out as follows.

\subsubsection{Environmental Configuration}
We generated 100 labyrinth layouts for evaluations. Each labyrinth is represented as an $8 \times 8$ grid consisting of pathways and obstacles generated using cellular automata. The start position is fixed at the bottom-left corner of the grid $(0, 0)$, and the goal position is set at the top-right corner $(7, 7)$. To maintain consistent complexity across experiments, we imposed a constraint on the number of obstacles: labyrinths with fewer than 8 or more than 16 obstacles were rejected. While the number of obstacles varies between 8 and 16, the optimal path length from the start to the goal remains fixed at 14 steps for every labyrinth. This shortest path is computed using Dijkstra's algorithm and corresponds to the Manhattan distance between two opposite corners of the $8 \times 8$ grid. This controlled complexity helps ensure a balanced level of challenge across all generated environments. The proportion of labyrinths categorised by obstacle count within the range of 8 to 16 is illustrated in Figure~\ref{fig:obstacle_count}, highlighting the constraint-imposed diversity in obstacle counts across the generated layouts, and ensuring varied yet controlled complexity in the evaluation environments.

\begin{figure*}[!t]
\centerline{\includegraphics[width=2.0\columnwidth]{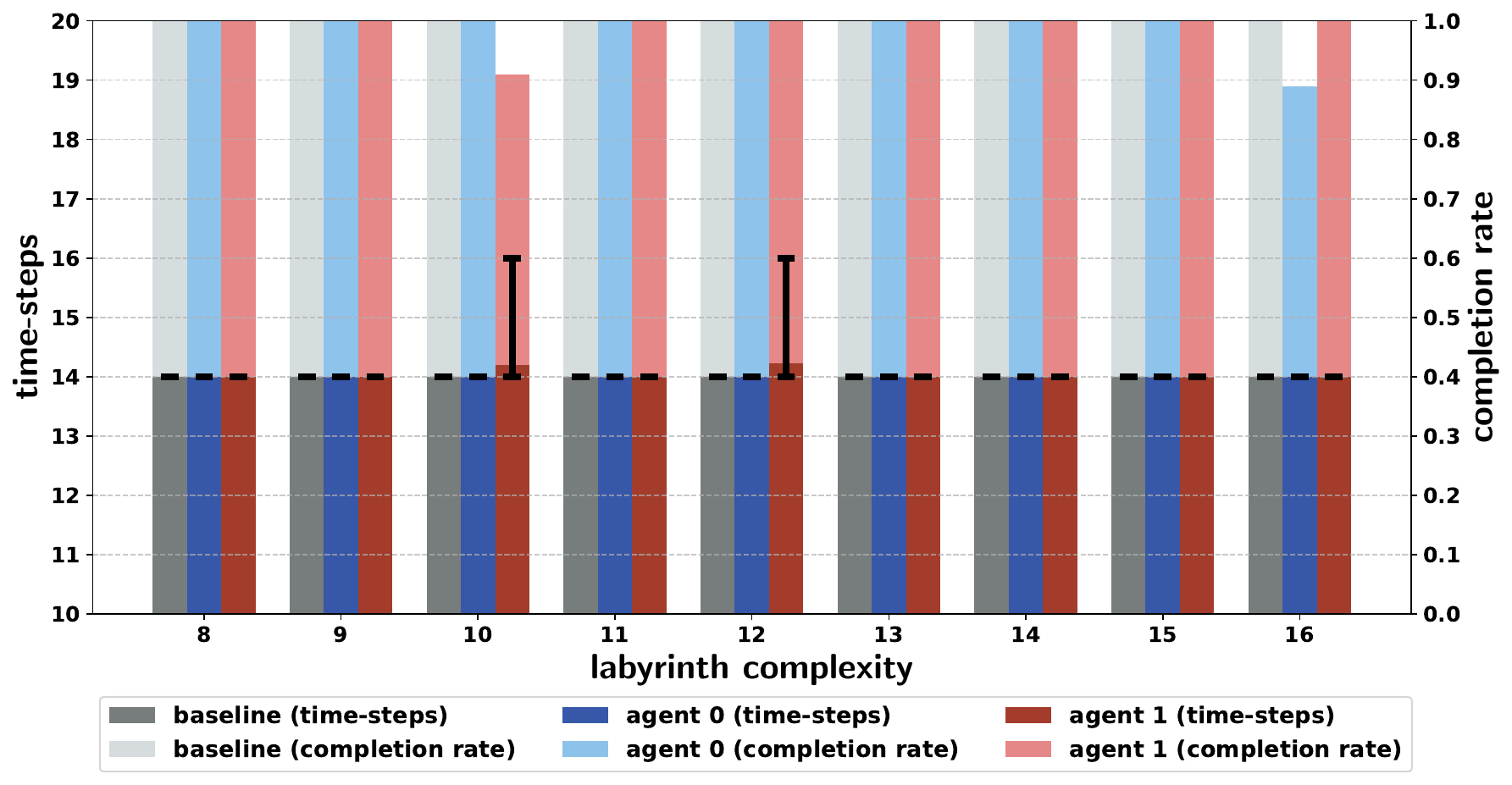}}
\caption{Agents' policy performance across labyrinths of varying complexity.}
\label{fig:game_performance}
\end{figure*}

\subsubsection{Feedback Configuration}
A small time penalty of $-0.04$ is applied at each step to motivate the agent to complete the task promptly. To encourage exploration and reduce redundancy, a revisit penalty of $-0.25$ is imposed whenever the agent revisits a previously explored grid cell. Collisions with obstacles incur a significant penalty of $-0.75$, incentivising the agent to learn effective obstacle avoidance. Upon successfully reaching the goal, the agent is awarded a terminal reward of $+1.00$, reinforcing the primary objective of solving the labyrinth efficiently. Additionally, a steganographic reward of $+1.00$ is granted if the agent’s episode is identifiable by the observer, encouraging the agent to generate distinguishable movement patterns while navigating the labyrinth.

\subsubsection{Agent's Model Configuration}
Each agent's deep Q-network was implemented using a CNN. The network takes as input a three-channel grid, where each channel represents a specific feature, such as the positions of obstacles, the agent and the goal. The network begins with two convolutional layers to extract spatial features from the input. The first convolutional layer applies 64 filters of size 3 with a stride of 1 and padding of 1, followed by an activation function to introduce non-linearity. The second convolutional layer applies 128 filters of the same kernel size, stride and padding, followed by another activation. The spatial features are flattened into a one-dimensional vector and passed through a fully connected linear layer, transitioning from spatial representations to actionable outputs that represent the Q-values for 4 possible actions.

\subsubsection{Observer's Model Configuration}
The observer model was implemented using an RNN. The input to the model is a sequence of three-channel grids, where each channel represents a specific feature at a given time step, such as the positions of obstacles, the agent and the goal. Initially, the input grids are flattened using a time-distributed layer, which processes each time step independently to prepare the data for sequential analysis. The flattened grid is then passed through a bidirectional LSTM network with an input size proportional to the flattened grid dimensions. The LSTM consists of two layers with 128 hidden units each and a dropout rate of 0.2 to prevent overfitting. The bidirectional LSTM enables the model to consider both past and future dependencies within the sequence. The final forward and backward hidden states from the LSTM are concatenated to form a feature vector for each sequence. This concatenated feature vector, which captures the dependencies of the agent’s movement patterns, is passed through a fully connected linear layer to produce the final output, representing the probability distribution over the agents' identities based on the observed trajectory.

\subsubsection{Learning Configuration}
Each agent's neural network and the observer's neural network were trained via back-propagation~\cite{Rumelhart:1986aa}, using their respective optimisers based on stochastic gradient descent with adaptive moment estimation~\cite{2015_Adam}. For each agent's model, the learning rate $\alpha$ was set to 0.001 to ensure gradual updates to the neural network weights during training, balancing convergence stability and speed. The discount factor $\gamma$ was set to 0.95, prioritising future rewards while moderately discounting their value, thus encouraging the agent to make decisions that optimise long-term cumulative rewards. For the observer's model, the learning rate was also set to 0.001. To avoid infinite loops during both learning and inference, a game-over threshold was introduced, limiting each trajectory to a maximum of 140 steps, corresponding to 10 times the optimal path length.

\begin{figure*}[!t]
\centerline{\includegraphics[width=2.0\columnwidth]{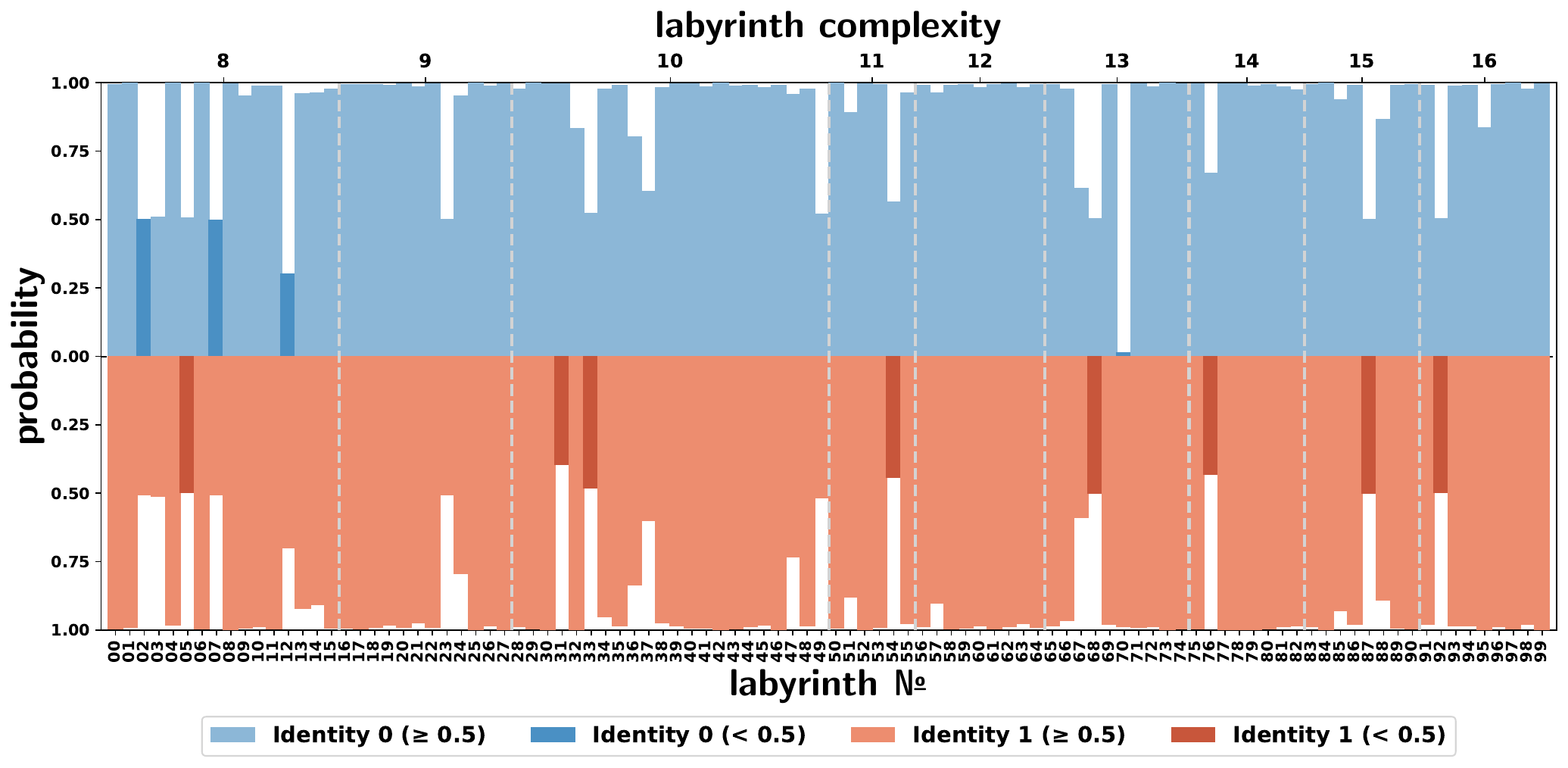}}
\caption{Observer's identification performance across labyrinth of varying complexity.}
\label{fig:communication_performance}
\end{figure*}

\subsection{Visualisation of Policies}
As visualised in Figure~\ref{fig:vis_maze}, one baseline agent and two steganographic agents were trained for each labyrinth, showcasing their learnt policies while navigating environments of varying complexity. The baseline agent, trained without steganographic feedback, followed a policy optimised solely for task completion. In contrast, the two steganographic agents embedded unique binary digits into their trajectories by incorporating steganographic feedback during training. The visualisations highlight how the steganographic agents balanced the dual objectives of encoding information and efficiently reaching the goal. Notably, in certain scenarios, such as the labyrinth with complexity 13, the cover trajectory and one of the stego trajectories were identical, demonstrating the indistinguishability of the steganographic channel. This behaviour affirms the ability of the steganographic agents to embed information covertly without compromising their primary navigation performance.

\begin{table}[!t]
\centering
\caption{Secrecy statistics under eavesdropper's steganalysis.}
\resizebox{\columnwidth}{!}{
\begin{tabular}{l|c|c}
\toprule
\midrule
\textbf{Metric} & \textbf{Learning Phase} & \textbf{Inference Phase} \\ 
\midrule
\textbf{Confusion Matrix}  & 
$\begin{bmatrix}
100 & 0 \\
4 & 96
\end{bmatrix}$ & 
$\begin{bmatrix}
50 & 50 \\
41 & 59
\end{bmatrix}$ \\ 
\midrule
\textbf{Precision}          & 1.0 & 0.541 \\ 
\midrule
\textbf{Recall}             & 0.96 & 0.59 \\ 
\midrule
\textbf{F1 Score}           & 0.9796 & 0.5646 \\ 
\midrule
\textbf{AUC-ROC}            & 0.9992 & 0.5641 \\ 
\midrule
\textbf{Accuracy (Cover)} & 1.0  & 0.5 \\ 
\midrule
\textbf{Accuracy (Stego)} & 0.96 & 0.59 \\
\midrule
\textbf{Overall Accuracy}   & 0.98 & 0.545 \\ 
\midrule
\bottomrule
\end{tabular}
}
\label{tab:secrecy}
\end{table}

\subsection{Evaluation of Distortion}
The performance of agents for solving labyrinths, reflecting the distortion caused by steganographic communication, is presented in Figure~\ref{fig:game_performance}, focusing on time-steps and completion rates. The time-steps (opaque/dark-coloured bars) represent the average number of steps required by each agent to reach the goal, calculated only for completed trajectories, with error whiskers indicating the range from minimum to maximum. The completion rates (translucent/light-coloured bars) represent the proportion of successful trials. The results demonstrate that the steganographic agents achieved performance nearly identical to the baseline agent, with negligible deviations in time-steps and a small proportion of incomplete trajectories. These findings suggest that embedding steganographic information minimally impacts navigation efficiency, allowing the steganographic agents to encode information into their trajectories without significantly distorting game performance.

\subsection{Evaluation of Capacity}
The performance of the observer for identifying the source of each trajectory, reflecting the capacity of the steganographic channel, is presented in Figure~\ref{fig:communication_performance}. The blue bars represent probabilities assigned to the identity of agent 0 for trajectories from agent 0, while the red bars indicate probabilities assigned to the identity of agent 1 for trajectories from agent 1. Misidentifications, defined as probabilities below 0.5 for the corresponding identity, are highlighted by darker shades of blue and red. The x-axis is segmented by labyrinth complexity, determined by the number of obstacles. The results reveal that the observer generally achieves high identification accuracy with occasional misidentifications, as indicated by most probabilities being greater than 0.5. This demonstrates the steganographic channel's capacity to reliably transmit one bit of information through each trajectory. These findings suggest that communicating secret information through a subliminal channel manifested in the form of game trajectories is a feasible and effective approach.

\begin{table*}[!t]
\caption{Robustness statistics under intruder's stochasticity.\label{tab:robustness}}
\centering
\resizebox{2.0\columnwidth}{!}{
\begin{tabular}{l | c c | c c c | c c c}
\toprule
\midrule
\multicolumn{1}{l}{environment} & \multicolumn{2}{|c}{cover trajectory (baseline)} & \multicolumn{3}{|c}{stego trajectory (agent 0)} & \multicolumn{3}{|c}{stego trajectory (agent 1)}\\
\midrule
stochasticity & \multicolumn{1}{|c}{completion} & \multicolumn{1}{c}{timesteps} & \multicolumn{1}{|c}{completion} & \multicolumn{1}{c}{timesteps} & \multicolumn{1}{c}{identifiability} & \multicolumn{1}{|c}{completion} & \multicolumn{1}{c}{timesteps} & \multicolumn{1}{c}{identifiability} \\
\midrule
0.0 & 100\%           & 14.00 & \phantom{0}99\% & 14.00 & 0.9596 & \phantom{0}98\% & 14.06 & 0.9286\\
0.1 & 100\% 		  & 15.99 & 100\% 			& 16.33 & 0.9300 & \phantom{0}98\% & 16.08 & 0.8980\\
0.2 & 100\% 		  & 21.48 & 100\% 			& 21.62 & 0.9500 & \phantom{0}99\% & 22.06 & 0.8788\\
0.3 & 100\% 		  & 26.19 & 100\% 			& 26.47 & 0.9300 & \phantom{0}99\% & 26.18 & 0.8990\\
0.4 & 100\% 		  & 26.30 & 100\% 			& 26.62 & 0.9100 & \phantom{0}99\% & 26.62 & 0.8990\\
0.5 & 100\% 		  & 26.65 & 100\% 			& 27.29 & 0.8800 & \phantom{0}99\% & 27.31 & 0.8384\\
0.6 & 100\% 		  & 31.10 & 100\% 			& 30.70 & 0.9100 & 100\%			   & 30.87 & 0.8400\\
0.7 & 100\% 		  & 31.95 & 100\% 			& 31.93 & 0.8500 & 100\%			   & 32.36 & 0.7900\\
0.8 & 100\% 		  & 35.70 & 100\% 		    & 35.68 & 0.8500 & \phantom{0}98\% & 36.50 & 0.7653\\
0.9 & \phantom{0}81\% & 55.20 & \phantom{0}78\% & 56.72 & 0.6410 & \phantom{0}82\% & 54.93 & 0.6220\\
\midrule
\bottomrule
\end{tabular}
}
\end{table*}

\subsection{Evaluation of Secrecy against Eavesdropper}
Table~\ref{tab:secrecy} presents the performance metrics of the steganalyser trained to distinguish between cover and stego trajectories across the same set of labyrinths. This experiment simulates an eavesdropper, Eve, operating under Kerckhoffs's principle, where the neural network architectures and learning framework of the steganographic agents are assumed to be publicly accessible. However, the collective configuration of random seeds and hyper-parameters, serving as the stego-key, remains secret. To reflect this setup, the agents used to generate training data for the steganalyser were arbitrarily initialised. The results demonstrate high detection accuracy during the learning phase, with statistical metrics nearing perfection (approximately 1.0). In contrast, inference statistics show a notable decline, approaching random guessing (approximately 0.5), indicating the failure of the steganalyser to generalise and reliably detect steganographic communication within unseen trajectories. These findings reveal the secrecy of action steganography, demonstrating that an eavesdropper cannot reliably breach the stego-system as long as the stego-key shared between Alice and Bob remains secret.

\subsection{Evaluation of Robustness against Intruder}
Table~\ref{tab:robustness} presents the performance metrics under varying levels of stochasticity, representing the probability that an agent is forced to deviate from its learnt policy to choose a random action. This experiment simulates an active intruder, Trudy, who introduces stochasticity to disrupt the agents' trajectories and undermine steganographic communication. The metrics include completion rate, average time-steps for completed trajectories, and identifiability, which measures the observer’s ability to correctly identify the source of a trajectory, facilitating secret communication between Alice and Bob. The results show that both cover and stego agents maintain high completion rates and acceptable time-steps even under very high stochasticity (up to 0.8). Identifiability remains close to or above 0.9 under moderate stochasticity (up to 0.4) and close to or above 0.8 under high stochasticity (up to 0.7). These findings highlight the robustness of action steganography, demonstrating that an intruder cannot significantly disrupt communication as long as Alice and Bob operate within controlled levels of environmental randomness.

\section{Conclusions}\label{sec:conclusions}
This study introduced action steganography, a novel paradigm for communicating hidden messages through the behavioural trajectories of multiple artificial intelligence agents within a shared environment. The proposed stego-system was validated through the game of labyrinth, with systematic evaluations across key metrics, including distortion, capacity, secrecy (against passive eavesdroppers) and robustness (against active intruders). These findings demonstrate the potential of artificial intelligence to enable dynamic steganographic media in the form of patterned trajectories. Future research could scale the framework to more dynamic environments involving a greater number of agents and more intricate interactions. We envision that the concept of action steganography will pave the way for expanding the universe of what constitutes steganographic media.

\newpage
\IEEEtriggeratref{29}
\bibliographystyle{Transactions-Bibliography/IEEEtran}
\bibliography{Transactions-Bibliography/bstcontrol, Bib/bib_actstego}

\end{document}